\documentclass[letterpaper, 10 pt, conference]{ieeeconf}  

\IEEEoverridecommandlockouts                              

\usepackage[english]{babel}

\usepackage{amsthm}
\usepackage{amsmath}
\usepackage{amssymb}
\usepackage{booktabs}
\usepackage[dvips]{graphicx}
\usepackage{multirow}
\usepackage{amsfonts}
\usepackage{enumerate}
\let\labelindent\relax
\usepackage{enumitem}
\usepackage{tabularx}
\usepackage{algorithm,algorithmic}
\usepackage{bm}
\usepackage{adjustbox}
\usepackage{xcolor}
\usepackage{siunitx}
\usepackage{mdframed}
\usepackage{adjustbox}
\usepackage{authblk}
\usepackage{color}
\usepackage{array}
\usepackage{tabularx}
\usepackage{xurl}
\usepackage{subcaption}
\usepackage{hhline} 
\usepackage{cite}
\makeatletter
\let\NAT@parse\undefined
\makeatother
\usepackage{relsize}
\usepackage{float}
\usepackage{pifont}
\usepackage{hyperref}

\usepackage{tabularx}
\usepackage{ragged2e}

\title{\LARGE \bf
Flying Together: Human-Guided Immersive Shared Control \\for Aerial
Robot Teams in Unknown Environments }
\author{Lou De Bel-Air$^{1,2}$, Luca Morando$^{1,3}$, Ruitao Chen$^1$, Keru Wang$^1$, Benjamin Jarvis$^2$, Charbel Toumieh$^2$,\\ Yang Zhou$^1$, Ken Perlin$^1$, Dario Floreano$^2$, and Giuseppe Loianno$^3$
\thanks{$^1$The authors are with New York University, NY 10012, USA.
{\tt\footnotesize
\{ld3280, luca.morando, rc4000, kw2727, yangzhou, kp1\}@nyu.edu}.}
\thanks{$^{2}$The authors are with the Ecole Polytechnique Federale de Lausanne, Station 9 CH-1015 Lausanne, Switzerland. {\tt\footnotesize email: \{benjamin.jarvis, charbel.toumieh,dario.floreano\}@epfl.ch.}}
\thanks{$^3$The author is with the University of California Berkeley,
Department of Electrical Engineering and Computer Sciences,
Berkeley, CA 94720, USA. {\tt\footnotesize email: loiannog@eecs.berkeley.edu}.}
\thanks{This work was supported by the NSF CPS Grant CNS-2603416, the NSF CAREER Award 2546659, and the DARPA YFA Grant D22AP00156-00.}
}

\begin{document}

\maketitle
\thispagestyle{empty}
\pagestyle{empty}

\begin{abstract}
While autonomous multi-robots can achieve safe and coordinated navigation, they often struggle to adapt to unforeseen conditions and to capture operator-driven objectives in unstructured environments. 
We present a Virtual Reality (VR)-based shared control framework for teams of drones operating in constrained and unknown environments, enabling real-time, user-guided exploration. At the core of our approach is a novel, user-guided motion-primitive-based planner that computes continuous, collision-free trajectories while continuously integrating operator input. This planner is coupled with an admittance controller, allowing the operator to flexibly influence team behavior and guide drones toward regions of interest that autonomous planners may overlook. The system supports mixed-reality operations with both physical and simulated drones, and implements a bilateral VR-based interface, allowing the operator to guide the robot team via migration points while receiving immediate visual feedback of the team state. Experimental results show that shared control improves obstacle avoidance, maintains inter-agent spacing, and reduces operator effort, demonstrating the feasibility and advantages of immersive, human-in-the-loop multi-robot navigation.
\end{abstract}


\section{Introduction}
\begin{figure*}[!t] 
  \centering
  \includegraphics[width=0.76\textwidth]{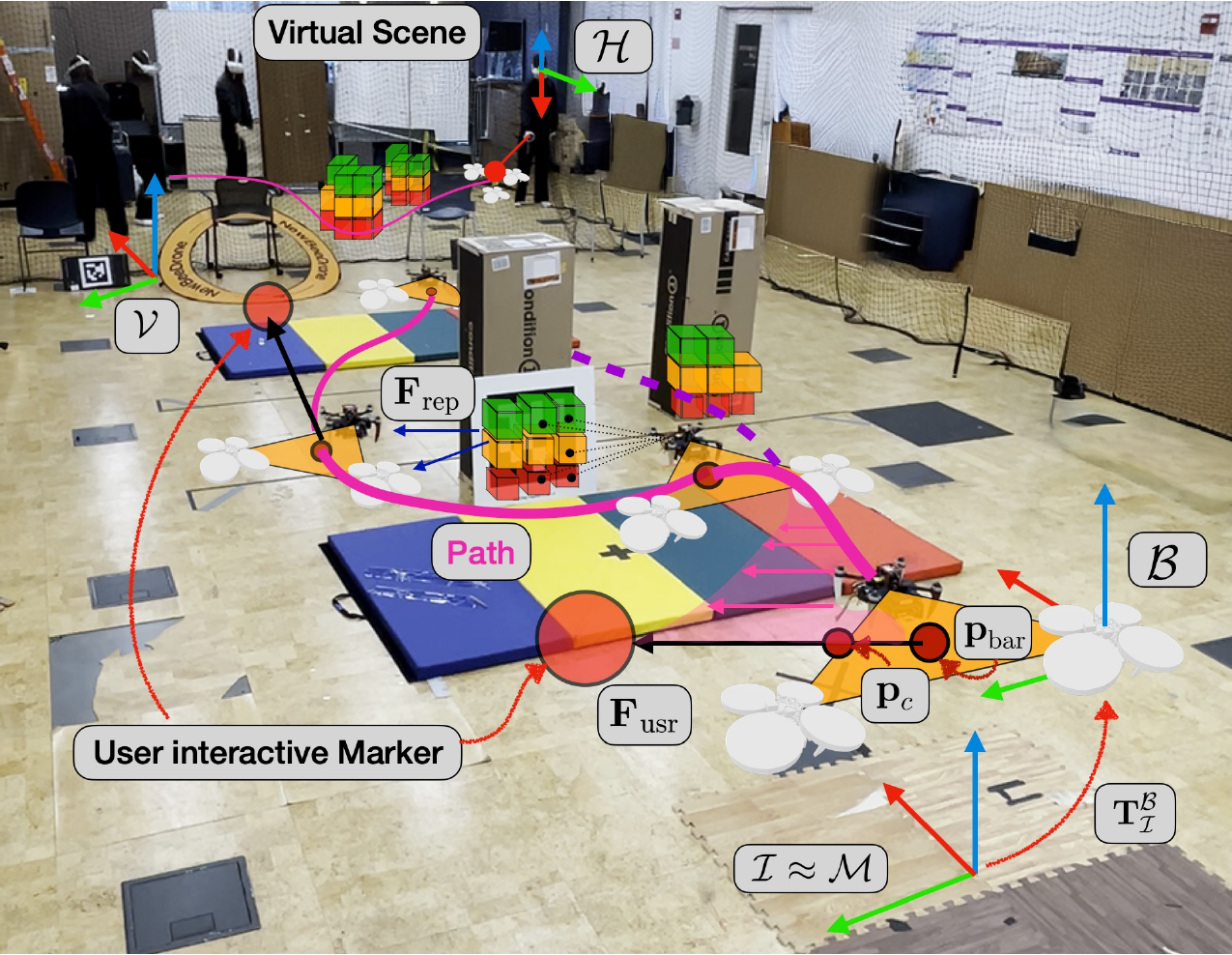} 
  \caption{Mixed-reality experiment with one real and two simulated drones (white). The operator in the background guides the drones in the VR environment using a red marker (\(\mathbf{F}_{\text{usr}}\)), and the planner reacts to this input in real-time (pink). \(\mathbf{p}_{\text{bar}}\) and \(\mathbf{p}_c\) are robots barycenter and migration point; \(\mathbf{F}_{\text{rep}}\) is the obstacle repulsion. Frames \(\mathcal{I}, \mathcal{M}, \mathcal{B}, \mathcal{V}, \mathcal{H}\) are defined in Section~\ref{sec:preliminaries}.
}
  \label{fig:intro}
  \vspace{-15pt}
\end{figure*}
 Unmanned Aerial Vehicles (UAVs) are increasingly deployed for tasks such as search and rescue~\cite{Michael_JFR}, environmental monitoring~\cite{Burschka2012}, and infrastructure inspection~\cite{Ozaslan2016}. In many of these applications, autonomous navigation offers scalability and can maximize coverage while reducing operator workload, yet full autonomy often limits user interaction and adaptability in complex, dynamic environments. Consider a disaster response scenario in which a swarm of drones explores a partially collapsed building. While autonomy can manage large-scale coverage, a human operator may still need to guide drones toward specific areas of interest, adjust their trajectories in real time, or prioritize regions based on visual cues unavailable to the planner. This tension between autonomy and user input motivates the use of shared control, a paradigm well established for manipulators but still limited for aerial robots given their virtually unbounded 3D workspace. In addition, the field still lacks an approach that scales to multi-robot coordination and incorporates real-time human guidance through immersive, virtualized interaction, which functions as an integral modality for operator input within the shared control loop.
 

The key problem we address in this work is how to leverage aerial robot teams for user-guided navigation and exploration in a way that is both immersive and goal-oriented, while still allowing the operator to influence trajectories for exploration purposes.
Recent advances have shown that aerial robot teams can outperform single-drone systems in exploration and coverage tasks due to their larger parallelized sensing and robustness to failures~\cite{quan2025}. At the same time, drone teleoperation research has demonstrated that 3D interfaces, such as Virtual Reality (VR), provide more intuitive control compared to traditional 2D interfaces, and even outperform First-Person-View (FPV) control in terms of task completion time, path optimality, and operator workload~\cite{Sanket}. Our prior work on user-in-the-loop drone navigation~\cite{Morando24} introduced an immersive framework for guided drone control. A subsequent user study~\cite{Morando25} demonstrated that operators placed high trust in mixed-reality (MR) representations and preferred guided trajectories over approaches relying on excessive autonomy. However, several limitations remained: (i) trajectories were only recomputed at intermediate goals, preventing escape from confined regions; (ii) the generated trajectories were not always dynamically feasible, limiting their applicability in multi-robot settings; and (iii) user control was minimal, with drones forced to strictly follow planner outputs without allowing operator adjustments. 

To address these challenges, we propose a 3D VR-based shared control framework for teams of aerial robots operating in unknown environments. At its core, we design a novel real-time, collision-free, minimum-time, operator-aware sampling-based motion planner. This planner, which balances autonomy with user-guided navigation, is integrated with an admittance control mechanism to ensure safe and dynamically feasible motion while maintaining flexibility for real-time operator inputs.
Motion primitives are used to guarantee feasibility at the team level, enabling consistent group behavior and collision-free coordination. Individual agent motion is coordinated by decomposing the overall group behavior and distributing it across the team, while accounting for cohesion, trajectory tracking, velocity consensus, and obstacle avoidance in a distributed fashion. Finally, we present a web-based immersive interface, an integral component of the shared control loop, which enables intuitive, real-time operator input and full control of aerial robot teams. While the web-based design introduces additional development complexity, it offers significant advantages: it is open-source, cross-platform, multi-user, and widely accessible, eliminating reliance on proprietary software and democratizing immersive human–robot interaction. This approach provides a transparent, reproducible, and widely deployable solution for multi-robot shared control.

An overview of the proposed system with one real and two virtual aerial robots, highlighting the immersive interface and planner adaptation, is shown in Fig.~\ref{fig:intro}.

\section{Related Works}

\label{sec:related_works}
Human–robot collaboration has progressed beyond industrial manipulators~\cite{Xing21} to include small-scale aerial robots, which are increasingly vital for search and rescue~\cite{Michael_JFR}, infrastructure inspection, and patrolling~\cite{Morando2020}. Interaction with aerial vehicles remains limited, often depending on visual feedback~\cite{Isop2019} or joystick control. Furthermore, current methods scale poorly when managing teams of aerial robots, as operators are typically restricted to assignment of high-level waypoints~\cite{Betancourt2022}.

\textbf{Shared Control}.
Shared control has often been studied in simpler systems such as manipulators~\cite{Stroppa2023_JIntellRobotSyst}. Extending these solutions to aerial robots is challenging due to their higher motion complexity, virtually unbounded workspace compared to traditional manipulators and the impossibility to have tactical feedback, which often requires a virtualization of the interaction. An approach for single agent is proposed in~\cite{Franceschini2023_ICUAS}, where operator commands are combined with autonomous behaviors to improve safety, precision, and task efficiency. Extending these approaches to team of aerial robots introduces unique challenges: multiple agents must coordinate in three-dimensional, unconstrained environments while simultaneously following human intent, maintaining inter-agent cohesion, and avoiding collisions. 

\textbf{Teleoperation}. 
Teleoperation remains one of the most direct approaches to robot control, yet turning robots into effective human partners rather than simple executors is still a challenge~\cite{Bentz2022}. Current autopilots can generate agile trajectories~\cite{Ales2022} or optimize exploration paths in indoor environments~\cite{mao2025}, but offer limited means for intuitive human-robot communication~\cite{Szafir2}. A promising step forward is made in~\cite{Morando24}, where an admittance controller based on interaction forces allowed the user to perceive feedback from both the robot and its environment. However, the planning support in that work was minimal and not scalable to multi-robot scenarios. Similarly, Jump Point Search has been proposed as a middle layer between user motion inputs and robot execution~\cite{Sanket,Harabor_Grastien_2014}, but these methods still lack the ability to generate real-time, dynamically feasible trajectories~\cite{liu2017}. Bridging this gap with planners that can seamlessly integrate user preferences and robot dynamics in cluttered environments remains an open research direction.

\textbf{Multi-Robot Exploration}. 
Several methods have been proposed for cooperative exploration using aerial and ground robots~\cite{toumieh2024high}. Although approaches rooted in traditional control theory can offer mathematical guarantees, they remain limited when handling complex indoor environments~\cite{Brambilla2013}, especially when a human operator is part of the control loop. This challenge is increasingly relevant as demand for aerial swarm systems grows, making Human-Swarm Interaction (HSI) a critical research area.

In most current multi-robot applications, operators are forced to divide their attention among individual robots increasing workload and reducing situational awareness issues that can have severe consequences~\cite{abdi2023safe,hocraffer2017meta}. Some promising solutions have emerged, including a swarm interaction framework grounded in cognitive models of human decision-making, designed to reduce operator load in complex tasks~\cite{ZHOU2025100029}, and a multi-robot task allocation mechanism that enables effective human oversight, as demonstrated in the DARPA Subterranean Challenge~\cite{Chen22}.

Despite these advances, a general and scalable framework for centralized, human-in-the-loop swarm control in constrained environments remains absent. 

\textbf{Immersive Interfaces for Multi-Robot Control}.
The advent of extended reality interfaces has significantly advanced intuitive and effective interaction in robotics, improving both information exchange and visualization~\cite{walker2022virtual}. These enable more natural communication of user intentions to robotic teammates~\cite{rosen2017communicating,coovert2014spatial}. In~\cite{Erat2018}, users interact with a virtual exocentric view of a drone and specify final poses via intuitive pick-and-place gestures, while in~\cite{angelopoulos2022drone}, operators assign sequences of waypoints. Similarly,~\cite{Szafir2} proposes a Mixed Reality (MR) framework that replaces the ground station with a holographic satellite map displaying live robot status updates, thereby improving situational awareness during human-robot collaboration.
More advanced approaches, such as~\cite{Sanket,Morando24}, extend this paradigm by introducing perception-based interaction, where robot information is directly translated into graphical cues facilitating cooperation. Meanwhile, recent progress in graphics, connectivity, and rendering quality has enabled new virtual environments like WebXR~\cite{WebXRspec}. These are customizable, web-based, cross-platform, and support multi-user interaction, showing promising results in robotics research and human-robot interaction tasks~\cite{Wang24,Wang24Perlin}.

Building on these advances, our approach integrates custom VR graphics with a low-latency communication layer linking the VR interface and the robots, combined with an operator-assisted multi-robot motion planner.

\section{Methodology}
\label{sec:Methodology}

\begin{figure*}[!t] 
  \centering
  \includegraphics[width=0.9\textwidth]{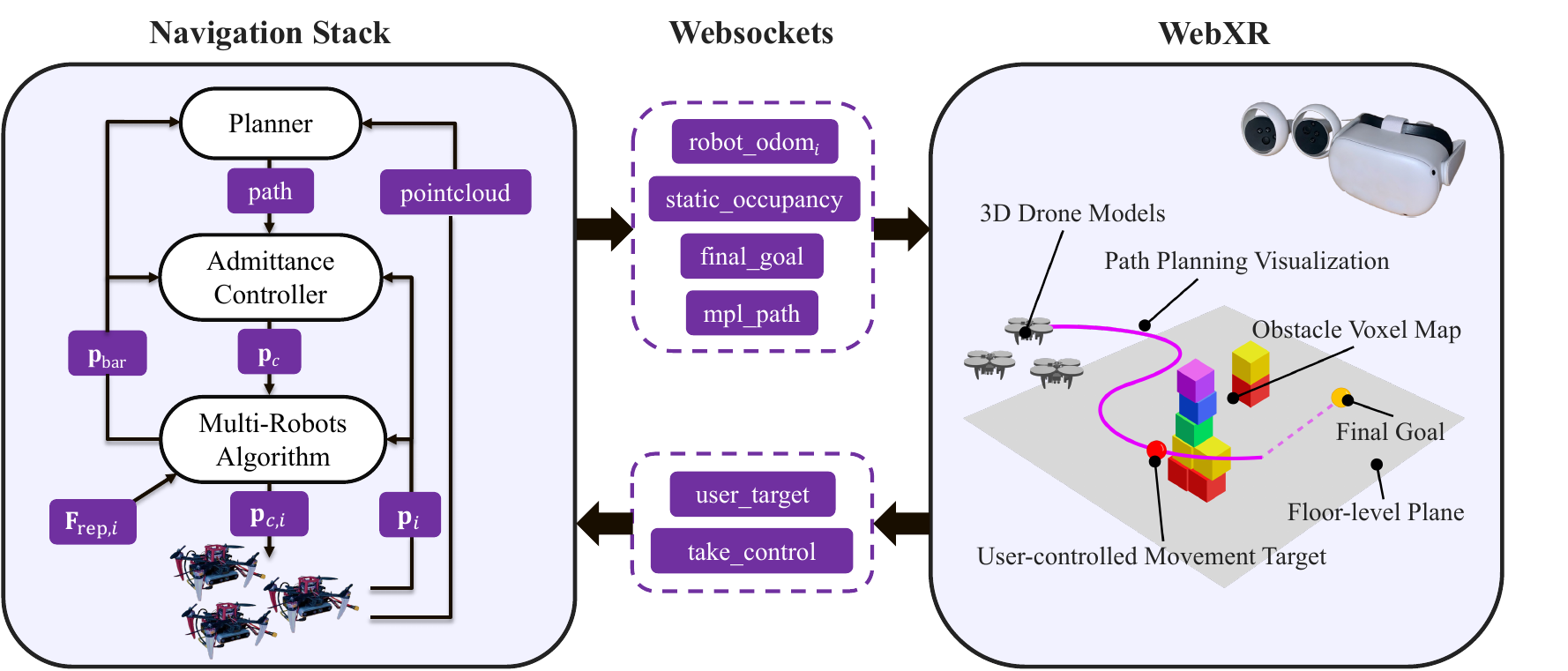} 
  \caption{Overview of the system: navigation stack (left; topics include barycenter of the robots \(\mathbf{p}_{\text{bar}}\), obstacle repulsive forces \(\mathbf{F}_{\text{rep},i}\), agent positions \(\mathbf{p}_i\), migration point \(\mathbf{p}_c\), and commanded positions \(\mathbf{p}_{c,i}\)), bidirectional WebSocket communication layer (center; topics detailed in Table I), and VR visualization of drones, local planner path, and environment (right).
}
  \label{fig:software_architecture}
  \vspace{-15pt}
\end{figure*}

In this section, we present the key components of the proposed immersive shared control framework, which allows operators and autonomous planners to collaboratively guide multi-robot teams. A motion-primitive-based planner (Section~\ref{sec:planner}) generates dynamically feasible paths, which are integrated with user input through an admittance controller (Section~\ref{sec:virtual-physical-interaction}). At the lower level, inter-agent force-based interactions provide robust multi-robot coordination (Section~\ref{sec:swarm}). Finally, Section~\ref{sec:webxr_platform} presents the virtual reality interface and low-latency middleware connecting the user to the robots. An overview of the complete framework is shown in Fig.~\ref{fig:software_architecture}.

\subsection{Frames, Notation, Localization, and Mapping}
\label{sec:preliminaries}
As shown in Fig.~\ref{fig:intro}, the inertial frame $\mathcal{I}$ is shared between the user and the robots, while the map frame $\mathcal{M}$ is aligned with $\mathcal{I}$ at the take-off location. Each robot $i$ has its own body frame $\mathcal{B}_i$. The user’s head-mounted display is represented by frame $\mathcal{H}$, and the interactive marker is attached to frame $\mathcal{K}$. Finally, the virtual world is represented by frame $\mathcal{V}$, which aligns with the inertial frame. The main frames are illustrated in Fig.~\ref{fig:intro}.
We denote by $\mathbf{T}_i^j$ the homogeneous transformation matrix from frame $i$ to frame $j$. 
Each robot $A_i^\mathcal{I}$, with $i \in \{0, \dots, N\}$, is localized in the inertial frame through the transformation $\mathbf{T}_\mathcal{I}^{\mathcal{B}_i}$, obtained from on-board state estimation and localization algorithms. The same alignment holds for the user, where $U^\mathcal{V}$ is expressed in the virtual frame $\mathcal{V}$, which is aligned with $\mathcal{I}$.


On the perception side, the agents can generate a point cloud $P_\mathcal{B}^\mathcal{I}$, initially expressed in its body frame $\mathcal{B}_i$ and then transformed into the map frame $\mathcal{M}$ using $\mathbf{T}_\mathcal{I}^{\mathcal{B}_i}$. This point cloud is converted into a real-time voxel map representation $V_{\mathcal{B}_i}^{\mathcal{I}}$, providing a geometric model of the environment. 

\subsection{User-in-the-Loop Motion Primitive Planner}
\label{sec:planner}


To assist the user-guided navigation, we employ a motion-primitive-based planner, which generates continuous, smooth trajectories by concatenating short-duration primitives derived from optimal control solutions~\cite{liu2017}. Unlike previous approaches for assisted drone teleoperation~\cite{Morando24}, which rely on geometrically disconnected RRT* segments, our framework ensures path continuity, minimizing abrupt maneuvers that could destabilize swarm formations. For real-time navigation in partially known environments, planning is performed within a 3D voxel map $V_{\mathcal{B}}^{\mathcal{I}}$ collected by one of the agents. This map is limited to the robot’s horizon $h$ and planning is performed toward a goal $G$ projected within this horizon. Precomputed motion primitives form a connected graph, where edges correspond to feasible state transitions and sequences are selected using graph search (A*) to minimize a cost function that balances smoothness, safety, and task efficiency
\begin{equation}
J^* = \min_{D,T}\; J_q(D) + \rho T + \rho_c J_c(D) + \rho_{\text{usr}} J_{\text{usr}}(D),
\label{eq:cost_fct}
\end{equation} 
where $D$ is a candidate trajectory, $T$ its duration, $J_q$ penalizes control effort, $J_c$ penalizes obstacle proximity, and $J_{\text{usr}}$ incorporates live user input. The weights $\rho$, $\rho_c$ and $\rho_{\text{usr}}$ adjust the relative influence of each term.

The key novelty of our approach is the user alignment term $J_{\text{usr}}$, which allows the input of the live operator to influence the planned trajectory in real time
\begin{equation}
J_{\text{usr}}(D) = \int_0^T \lVert \mathbf{F}_{\text{usr}} \rVert
\bigl(1 - \hat{\mathbf{F}}_{\text{usr}} \cdot \hat{\mathbf{v}}_D(t) \bigr) e^{-t/\tau} dt.
\end{equation}
Here, the user force $\mathbf{F}_{\text{usr}}$ encodes the user’s intent and is computed as $\mathbf{F}_{\text{usr}} = \mathbf{K}_{p}(\mathbf{p}_{\text{bar}} - \mathbf{p}_{u}) - \mathbf{K}_{d}(\dot{\mathbf{p}}_{\text{bar}}- \dot{\mathbf{p}}_{u})$,
with $\mathbf{K}_{p}$ and $\mathbf{K}_{d}$ the proportional and derivative gains, $\mathbf{p}_{\text{bar}}$ the barycenter (mean position of agents), and $\mathbf{p}_{u}$ the position where the user intends to drag the robots. 
In $J_{\text{usr}}$, the magnitude $\lVert \mathbf{F}_{\text{usr}} \rVert$ ensures that stronger inputs exert proportionally greater influence on the trajectory. The unit vector $\hat{\mathbf{F}}_{\text{usr}}$ and $\hat{\mathbf{v}}_D(t)$, which is the unit velocity vector along the primitives, favor trajectories aligned with the user’s intention while penalizing misaligned ones. The exponential decay $e^{-t/\tau}$ restricts user influence to the early portion of the trajectory, allowing later primitives to prioritize efficiency toward the goal. The effect is shown in Fig.~\ref{fig:intro}, where the optimal trajectory is smoothly perturbed to reflect user intention.

During execution, the planner incrementally sequences motion primitives into a locally feasible trajectory. This process is re-evaluated every two seconds, enabling adaptation to dynamic obstacles while maintaining continuous user involvement.

\subsection{Variable Admittance Controller}
\label{sec:virtual-physical-interaction}
Once a path has been generated by the planner, the challenge is that the drones do not follow it autonomously and instead adapt to the user’s commands. To address this, we define an extended Variable Admittance Controller (VAC)~\cite{Morando24}, which combines the motion-primitive path with the user’s intent to generate a commanded position for the agents to follow, while maintaining proximity to the planned trajectory.
The virtual force input driving the VAC is defined as $\mathbf{F}_v = \mathbf{F}_{\text{usr}} + \mathbf{F}_{\text{rep}}$, where $\mathbf{F}_{\text{usr}}$ is the user force introduced in Section~\ref{sec:planner}, and $\mathbf{F}_{\text{rep}}$ is an obstacle repulsion force that ensures the commanded position remains clear of obstacles. The repulsion force is computed as
\begin{equation}
\scalebox{0.97}{$\displaystyle
\mathbf{F}_{\text{rep}} = \sum_{i=0}^{N_v} \mathbf{F}_{r_i}
= \sum_{i=0}^{N_v} \frac{F_s}{k} e^{-\lambda d_i} \left( 1 - e^{h - d_i} \right)
$}
\end{equation}
with $v_i \in V^W$ the voxels within the robot horizon $h$, $d_i$ is the distance to voxel $v_i$, $F_s$ is the maximum force, and $k = 1 - e^h$ normalizes the force  $\|\mathbf{F}_{r_i}\| = F_s$ when $d_i = 0$.

Next, the barycenter of the agents, $\mathbf{p}_{\text{bar}}$, is projected onto the path generated by the planner to define a sequence of reference positions $\mathbf{p}_r$. The VAC then computes a commanded position $\mathbf{p}_c$, which serves as the migration point for the team and evolves according to mass–spring–damper dynamics
\begin{equation}
\scalebox{0.97}{$\displaystyle
\mathbf{M} \left(\ddot{\mathbf{p}}_c - \ddot{\mathbf{p}}_r\right)
+ \mathbf{D}\left(\dot{\mathbf{p}}_c - \dot{\mathbf{p}}_r\right)
+ \mathbf{K}\left(\mathbf{p}_c - \mathbf{p}_r\right)
= \mathbf{F}_v
$}
\label{eq:admittance1}
\end{equation}
where $\mathbf{M}$, $\mathbf{D}$, and $\mathbf{K}$ are diagonal matrices encoding mass, damping, and stiffness in the world frame. 

\subsection{Multi-robots Coordination Execution}
\label{sec:swarm}
The commanded position $\mathbf{p}_c$ produced by the admittance controller must be distributed among the individual agents to enable safe and coordinated tracking.
We adopt a force based algorithm inspired by Olfati–Saber~\cite{Olfati-Saber2006_TAC}.  For each agent $i$, the reference acceleration is computed as the combination of inter-agent cohesion, velocity consensus, centralized tracking, and obstacle repulsion
\begin{equation}
\scalebox{0.95}{$
\begin{aligned}
\mathbf{a}_i^{\mathrm{ref}} =& 
\underbrace{\sum_{j\in\mathcal{N}_i} \boldsymbol{\phi}(\mathbf{p}_{ji})}_{\text{cohesion}}
+ \underbrace{\beta \sum_{j\in\mathcal{N}_i} (\mathbf{v}_j-\mathbf{v}_i)}_{\text{consensus}} \\
&+ \underbrace{K_p(\mathbf{p}_c-\mathbf{p}_i) + K_v(\mathbf{v}_c-\mathbf{v}_i)}_{\text{tracking}}
+ \underbrace{\mathbf{F}_{\text{rep},i}}_{\text{obstacle}} 
\end{aligned}
$}
\end{equation}
where $\mathbf{p}_{ji} = \mathbf{p}_j - \mathbf{p}_i$, with $\mathbf{p}_i$ and $\mathbf{v}_i$ the position and velocity of agent $i$, and $\mathbf{F}_{\text{rep},i}$ the obstacle repulsion force described in Section~\ref{sec:virtual-physical-interaction}. This repulsion term must also be incorporated into the multi-robots algorithm itself, as the admittance controller ensures that the migration point avoids obstacles, but each agent still needs to individually avoid collisions.
The inter-agent cohesion function $\boldsymbol{\phi}(\mathbf{p}_{ji})$ is defined as
\begin{equation}
\scalebox{0.95}{$
\boldsymbol{\phi}(\mathbf{p}_{ji}) =
\alpha \, \rho_h\left(\frac{\sigma(\|\mathbf{p}_{ji}\|)}{\sigma(R)}\right)
\, \bigl(\sigma(\|\mathbf{p}_{ji}\|) - \sigma(d_{ij}^{\text{ref}})\bigr)
\, \frac{\mathbf{p}_{ji}}{\|\mathbf{p}_{ji}\|}
$}
\end{equation}
where $\sigma(\cdot)$ denotes the $\sigma$-norm, $\rho_h(\cdot)$ is a smooth cutoff function, $R$ is the interaction radius, and $d_{ij}^\text{ref}$ is the desired inter-agent distance.  

Acceleration is converted into a position command $\mathbf{p}_{c,i}$ for each $i^{\text{th}}$ agent  through double integration. Both acceleration and velocity are saturated to respect the dynamic limits of the drones and prevent unsafe maneuvers.

\subsection{Virtual Reality Platform and Communication Interface}
\label{sec:webxr_platform}
By providing an immersive 3D interface of both the environment and the robot team, VR naturally functions as an integral modality for operator input within the shared control loop, enhancing operator situational awareness, enabling intuitive manipulation of migration points, and supporting real-time guidance that complements the motion-primitive planner and admittance controller.

To create an accessible and versatile immersive interface, we develop a virtual reality framework based on WebXR\footnote{\url{https://www.w3.org/TR/webxr/}}. WebXR is a web-based framework that enables immersive Virtual and Augmented Reality experiences directly within web browsers, supporting both desktop and mobile head-mounted displays. It provides a flexible environment for rendering 3D content, capturing user interactions, and displaying spatial information in real time. WebXR also allows multiple users to share a virtual space and interact with dynamic objects, making it a natural platform for human-robot interaction and control.

In the context of aerial multi-robots teleoperation, WebXR provides the operator with a real-time visualization of the robots, trajectories and environmental context (Fig.~\ref{fig:software_architecture}). To seamlessly synchronize this virtual representation with the physical robots, a dedicated UDP WebSocket-based middleware is implemented. It bridges the control and perception stack running on the robots with the WebXR server running on the headset, enabling real-time communication of key visual cues, such as robot odometry, planned trajectories, and 3D map. Through this interface, the operator can monitor and influence the team in an intuitive, low-latency manner, while maintaining a consistent and interactive shared representation between the virtual and physical worlds.
The implemented topics and related messages in the custom WebSocket based communication middleware are both visualized in Fig.~\ref{fig:software_architecture} and summarized in Table \ref{tab:middleware_topics}.

\begin{table}[!htbp]
    \centering
    \small 
    \caption{WebSocket middleware topics}
    \label{tab:middleware_topics}
    \begin{tabularx}{\columnwidth}{@{}l>{\RaggedRight}X>{\RaggedRight}X@{}}
        \toprule
        \textbf{Topic Name} & \textbf{Message Type} & \textbf{Data Transmitted} \\
        \midrule
        robot$\_$odom$_i$ & webxr\_robot\_pos & Robots' odometry\\
        \addlinespace
        static\_occupancy & webxr\_static\_occupancy & Point cloud representing the 3D occupancy map\\
        \addlinespace
        final\_goal & webxr\_final\_goal& High level goal in the map\\       
        \addlinespace
        mpl\_path & webxr\_mpl\_path & Path computed by the motion primitive local planner  \\
        \addlinespace
        take\_control & webxr\_take\_control & Boolean flag to switch control of the robots \\
        \addlinespace
        user\_target & webxr\_des\_pos & The user's desired position and orientation \\
        \bottomrule
    \end{tabularx}
    \vspace{-20pt}
\end{table}

\section{Results}
\label{sec:results}
We validate the proposed framework using a team of quadrotors (one physical quadrotor and two simulated agents). The experiments evaluate the impact of the user
alignment term in the motion primitive planner and its role
in supporting immersive, user-in-the-loop multi-robot navigation.
Two scenarios are compared: (i) a baseline case where the
user alignment term $J_{\text{usr}}$ is disabled, enforcing shortest-path
planning; and (ii) a shared-control case where $J_{\text{usr}}$ is active,
allowing the operator to influence the trajectory. In both conditions, the user interacts with the drones through the VR interface by moving continuously a migration point marker, while the team maintains cohesion using the coordination strategy described in Section~\ref{sec:swarm}.

\begin{figure*}[!t] 
  \centering
  \setlength{\abovecaptionskip}{-1.5pt} 
  \includegraphics[width=\textwidth]{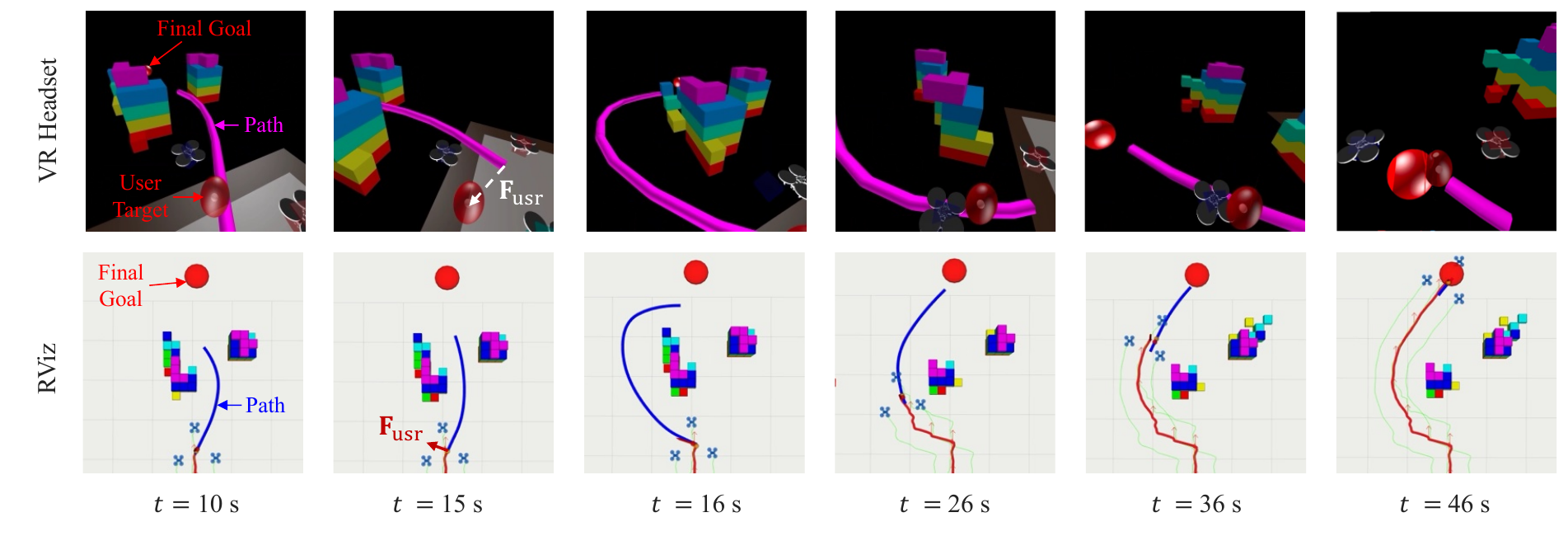} 
  \caption{Key frames from the shared-control experiment, showing the first-person VR view and the corresponding RViz visualization of the motion primitive planner's evolution as the user interacts with the VR target to guide the aerial team.}
  \label{fig3:results_frames}
  \vspace{-15pt}
\end{figure*}

\subsection{Robotic Platform and VR Interface} 

The real drone used in the experiments is a quadrotor with a total weight of $1.508~\si{kg}$ and thrust-to-weight ratio of $4\!:\!1$.
The platform is equipped with a PixRacer Pro flight controller and an NVIDIA Jetson Orin as the central processing unit, running Ubuntu 22.04 and ROS~2\footnote{\url{www.ros.org}}.
For the perception side, the drone leverages an Intel Realsense D455 stereo camera for localization using a customized version of OpenVINS~\cite{Geneva2020_ICRA}, which estimates the robot state at $100~\si{Hz}$.
A neural depth package~\cite{saviolo2025} is adopted for mapping purposes, providing a clear depth image as input to the NvBlox-based mapping package \cite{Millane2024} from which a point cloud $P_\mathcal{B}^\mathcal{I}$ is generated. 
The line of sight of the robot perception is set at $3~\si{m}$ with a voxel resolution of $20~\si{cm}$.
The team is composed of two additional simulated robots, with their dynamics computed onboard the real quadrotor. 
For the VR interface, we employ the Meta Quest 2 headset running Meta Horizon OS, a public-level headset suitable for research prototyping.

The VR environment is rendered with layered feedback: drones as 3D models, ground plane, planned trajectories as 3D strokes, and obstacles as a voxel map reconstructed online from occupied-space samples. These samples are then rendered in three.js\footnote{\url{https://threejs.org}} using InstancedMesh, a technique that draws thousands of identical objects (here cubes) in a single GPU call, enabling smooth real-time updates. Voxel colors encode height with a red–green–purple gradient from low to high, helping users to quickly interpret the environment.

WebXR is a server-based, cross-user, and cross-platform VR framework that runs in the browser, transmitting only the visual cues required for headset rendering. To minimize the latency with the server, we adopt a Netgear NightHawk V2 router, providing a minimal latency between the robot, the VR server and the headset, with a measured ping of only $5-8~\si{ms}$. 
Communication relies on both ROS~2 topics using FastRTPS middleware and custom WebSocket messages (Section \ref{sec:webxr_platform}).

The flying experiments are conducted in an indoor flying arena of approximately $60~\mathrm{m}^2$. 
The operator performs VR control in a separate sub-area of the same space, ensuring real-time interaction with the robot team while maintaining a safe separation from the real quadrotor (see Fig.~\ref{fig:intro}).

\subsection{Experimental Parameters}

\textit{Motion Primitive Planner.} The planning horizon is set at the line of sight of the robot perception to $h = 3~\mathrm{m}$ with a motion primitive duration of $\Delta t = 1.3~\mathrm{s}$. The cost function weights are chosen to balance trajectory duration ($\rho = 1.5$), obstacle force field ($\rho_c= 0.05$), and user influence ($\rho_{\text{usr}} = 35$) with an exponential decay constant $\tau = 0.8$. The goal is defined at $G^{\mathcal{I}} = [5.5, -0.1]~\mathrm{m}$, with a tolerance of $0.5~\mathrm{m}$. To address the continuous exploration and the missing knowledge of the map a priori, a virtual goal $G_v^I$ is projected on the robot horizon $h$, resulting as local destination of the planner motion primitives. 
 
\textit{Admittance Controller.} We follow the approach from~\cite{Morando24} to set the parameters, modifying only the obstacle force field $\mathbf{F}_{\text{rep}}$ to use an exponential decay ($\lambda = 0.55$) with a maximum magnitude of $F_s = 25.0~\mathrm{N}$ for smooth obstacle avoidance.

\textit{Team Coordination:} The desired inter-agent distance is $d_{ij}^\text{ref} = 1.5~\si{m}$. To keep the formation while allowing flexibility near obstacles, cohesion, alignment, and tracking gains are set to $\alpha = 1.0~\si{s^{-2}}$, $\beta = 0.3~\si{s^{-1}}$, and $K_p = 2.0~\si{s^{-2}}$, with damping gain $K_v = 1.5~\si{s^{-1}}$. Agents interact within a radius $R = 5~\si{m}$, with $\sigma$-norm smoothing $\epsilon = 0.08$ and a cutoff transition $h = 0.1$ for smooth decay of interactions.

 \begin{figure*}[!t] 
  \centering
  \includegraphics[width=0.9\textwidth]{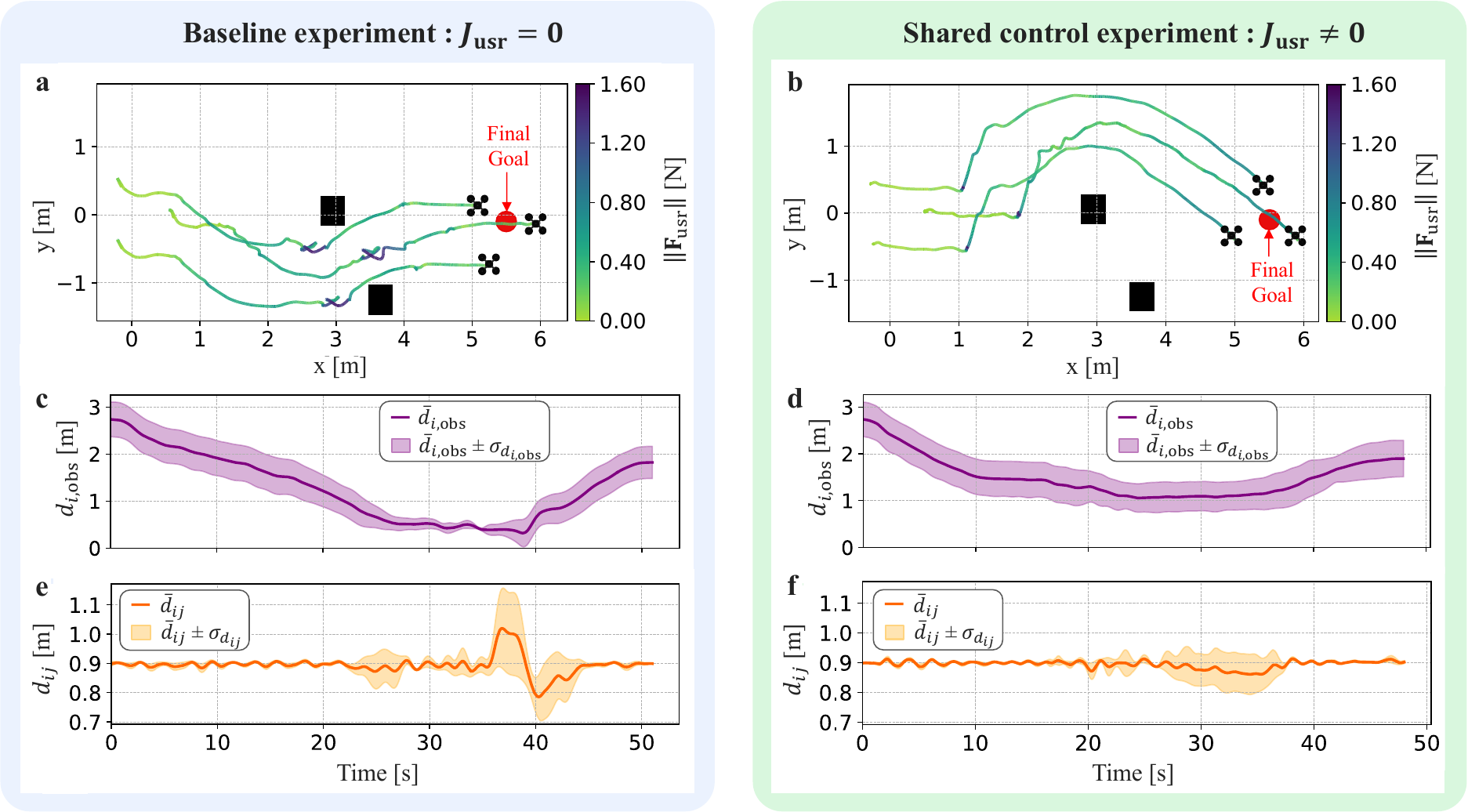} 
  \caption{Drone trajectories and multi-robots metrics for baseline and shared-control experiments. 
(a,b) Trajectories colored by user force $\mathbf{F}_{\text{usr}}$. 
(c,d) Distance to obstacles $d_{i,\text{obs}}$. 
(e,f) Inter-agent distances $d_{ij}$. 
Smoothed over 50 samples.} 
  \label{fig4:results_forces}
  \vspace{-15pt}
\end{figure*}

\subsection{Experimental Setup and Interaction}

To validate our approach in realistic human–drone interactions within unknown constrained environments, we designed the following experiments.
As described above, the team consists of one real robot $A_{0,r}$, and two simulated robots $A_{1,s}$  and $A_{2,s}$. 
While the simulated robots are visible to the user through the virtual environment, the real robot can perceive their presence through a continuous interaction based on cohesion forces and managed by the coordination control algorithm described in Section~\ref{sec:Methodology}. 
All robots, real or simulated, are subject to the user pulling force $\mathbf{F}_{\text{usr}}$ and the obstacle forces $\mathbf{F}_{\text{rep}}$ perceived in the real world through the agent $A_{0,r}$. 

The user guides the drone team in VR using the handheld controller, while perceiving the environment correctly localized via the transformation $\mathbf{T}_{\mathcal{V}}^{\mathcal{I}}$ between the virtual ($\mathcal{V}$) and inertial ($\mathcal{I}$) frames (Fig.~\ref{fig:intro}).
After activating the WebSocket connection, the user can direct the team by dragging a virtual marker from the barycenter, indicating the target position for the team. This input goes through the admittance controller to update the team’s migration point $\mathbf{p}_{c}$, allowing the drones to track it in real time.

For visualization, we present experiments with the obstacle configuration shown in Fig. \ref{fig3:results_frames}. However, the setup is not restricted to this case, as the algorithm adapts to different scenarios, making the experiments repeatable under various environmental conditions.

\subsection{Trajectory Analysis and Team Performance}
Fig.~\ref{fig3:results_frames} shows representative frames of the shared-control experiment, including the operator's first-person headset view and the corresponding RViz visualization (in ROS). In VR, the path computed by the motion primitive planner is displayed in pink, while the operator interacts with the red target to provide the force $\mathbf{F}_{\text{usr}}$ to the admittance controller. In the first frame ($t=10$~\si{s}) the user target is not moving, leading to $J_{\text{usr}} \approx 0$, and the displayed path corresponds to the shortest trajectory, passing through the two obstacles. At $t=15$ \si{s}, the user moves the target away from the precomputed path to suggest a safer path that avoids both obstacles. During this interval, the robots remain close to the precomputed path, as the admittance controller prevents large deviations. One second later, once the path is recomputed, the planner incorporates $\mathbf{F}_{\text{usr}}$ into the cost function and generates a new path aligned with the user's intended direction, as shown in the third frame. From this point, the operator continuously moves the target along the recomputed path until the team reaches the goal. In the final stages, the user no longer deviates the target from the path, and the planner repeatedly outputs the shortest path to the goal. 
\begin{table}[b]
\vspace{-20pt}
\centering
\caption{Multi-robots performance metrics for baseline and shared-control experiments}
\label{tab:metrics}
\small
\begin{tabularx}{\linewidth}{lcc} 
\toprule
\textbf{Metric} & \textbf{Baseline} & \textbf{Shared-Control} \\
\midrule
Average distance traveled [\si{m}] & \textbf{6.8} & 6.9 \\
Time to reach the goal [\si{s}] & 51 & \textbf{48} \\
Mean velocity [\si{m/s}] & 0.133 & \textbf{0.144} \\
Min. distance to obstacles [\si{m}] & 0.07 & \textbf{0.67} \\
Min. distance inter-agent [\si{m}] & 0.68 & \textbf{0.77} \\
Average user force [\si{N}]  & \textbf{0.38} & 0.39 \\
\bottomrule
\end{tabularx}
\end{table}
For both conditions, drone trajectories, distances to obstacles, and inter-agent distances are summarized in Fig.~\ref{fig4:results_forces}. In the baseline, $\mathbf{F}_{\text{usr}}$ is active to let the user drag the drones along the computed path, but the cost $J_{\text{usr}}$ is disabled in the planning algorithm, preventing trajectory modification and forcing the robots to stay on the shortest path through the obstacles. This requires the operator to apply larger pulls on the migration point, inducing oscillations between the obstacles and increasing $\mathbf{F}_{\text{usr}}$ during 9~\si{s}. As the team passes through the gap, the distance to the closest obstacle $d_{i,\text{obs}}$ drops to 0.07~\si{m}, causing large fluctuations in inter-agent distances and higher collision risk. In the shared-control experiment, a short peak in $\mathbf{F}_{\text{usr}}$ occurs when the operator initially deviates the path, but the required force quickly stabilizes. The drones avoid both obstacles while maintaining nearly constant inter-agent distances, obstacle clearances, and user force. These results fully validate the operator-assisted planning approach, enabling trajectories that are impossible under standard planning and demonstrating a more stable, intuitive control process while maintaining assisted navigation at all times.
These patterns are confirmed in the performance metrics reported in Table~\ref{tab:metrics}. While the traveled distance remains nearly unchanged between conditions, the team reaches the goal about 6\% faster under shared control. Most importantly, obstacle clearance improves substantially (0.67~\si{m} vs. 0.07~\si{m} in the baseline), and inter-agent spacing increases by $13\%$. 
\subsection{Discussion}
The results demonstrate that incorporating the user alignment term $J_{\text{usr}}$ in the motion primitive planner effectively enhances mixed-reality multi-robot teleoperation. Shared control enables the operator to continuously influence the drones trajectory, resulting in safer and more coherent formations without increasing operator effort.

The experiments also validate the different modules of the proposed framework for VR-based multi-robot teleoperation, combining user-in-the-loop motion-primitive planner (Section \ref{sec:planner}), admittance controller (Section \ref{sec:virtual-physical-interaction}), Olfati–Saber coordination strategy (Section \ref{sec:swarm}), and bilateral interaction with the VR framework (Section \ref{sec:webxr_platform}). 

Our framework preserves operational assistance even when the operator deviates from precomputed paths, a feature absent in prior works such as~\cite{Morando24}, while simultaneously ensuring safety and flexibility.



\section{Conclusion}
\label{sec:conclusion}
We presented a VR-based shared-control framework for teleoperating a team of drones in unknown environments. By combining a motion-primitive planner with an admittance controller, the system generates continuous, collision-free trajectories while allowing real-time operator-directed modifications. Mixed-reality experiments with both physical and simulated drones demonstrate that leveraging user intuition can concurrently improve obstacle clearance, maintain inter-agent distances, and reduce operator effort compared to fully autonomous planning.

This work provides a proof-of-concept for immersive human-in-the-loop multi-drone teleoperation, emphasizing the benefits of operator flexibility, intuitive guidance, and bilateral interaction. Future works will investigate scaling to a larger number of robots, integrating more advanced swarm algorithms, and conducting user case studies to quantify operator performance and system usability.




\bibliographystyle{IEEEtran}
\bibliography{reference}

\end{document}